# Multi-Stream Networks and Ground-Truth Generation for Crowd Counting

**Rodolfo Quispe**

University of Campinas, Institute of Computing
Av. Albert Einstein 1251, Campinas, SP - Brazil, 13083-852
e-mail: quispe@liv.ic.unicamp.br

**Darwin Ttito**

University of Campinas, Institute of Computing
Av. Albert Einstein 1251, Campinas, SP - Brazil, 13083-852
e-mail: ttito@liv.ic.unicamp.br

**Adín Ramírez Rivera**

University of Campinas, Institute of Computing
Av. Albert Einstein 1251, Campinas, SP - Brazil, 13083-852
e-mail: adin@ic.unicamp.br

**Helio Pedrini**

University of Campinas, Institute of Computing
Av. Albert Einstein 1251, Campinas, SP - Brazil, 13083-852
e-mail: helio@ic.unicamp.br

***Abstract*** *– Crowd scene analysis has received a lot of attention recently due to the wide variety of applications, for instance, forensic science, urban planning, surveillance and security. In this context, a challenging task is known as crowd counting [1–6], whose main purpose is to estimate the number of people present in a single image. A Multi-Stream Convolutional Neural Network is developed and evaluated in this work, which receives an image as input and produces a density map that represents the spatial distribution of people in an end-to-end fashion. In order to address complex crowd counting issues, such as extremely unconstrained scale and perspective changes, the network architecture utilizes receptive fields with different size filters for each stream. In addition, we investigate the influence of the two most common fashions on the generation of ground truths and propose a hybrid method based on tiny face detection and scale interpolation. Experiments conducted on two challenging datasets, UCF-CC-50 and ShanghaiTech, demonstrate that using our ground truth generation methods achieves superior results.*

## 1. INTRODUCTION

The task of crowd counting aims to estimate the number of people from a single RGB (Red-Green-Blue) image. The problem has a significant impact on several applications, for instance, urban planning, forensic science, surveillance and security [2, 7–10]. The main challenge in this task is the aggressive variation in scale and perspective of people in the images. Therefore, it can be complicated to differentiate between background and people (Figure 1).

Initial approaches used more classical people detection algorithms to directly count people in the image. For instance, Idrees et al. [16] proposed to obtain a headcount by mixing several features. They used a combination of head detection based on histogram of oriented gradient, handcrafted Fourier analysis, and interest-point based counting, then processed the resulting features with multi-scale Markov random field. Similar to other tasks in computer vision, handcrafted features often suffer from a decrease in accuracy when subjected to heavy variation in illumination, scale, severe occlusion, perspective and distortion.

To overcome the limitations of handcrafted methods, the seminal work of Zhang et al. [17] proposed a Multiple Stream Neural Network (MSNN) to estimate density maps. A density map represents the spatial distribution of people in an image, and it is more suitable for real-life applications, since it gives a notion of the people spatial distribution. The popularity of density maps has grown in deep learning methods [3, 11–16], such that they have become the default option for prediction of deep networks [3].

The main idea behind the MSNN is to specialize each stream at a specific person's scale. Thus, each stream follows similar architecture, however, with different filter sizes. Therefore, the active field of each stream differs according to the scale on which they focus. Following the work developed by Zhang et al. [17], many other MSNN variations have been proposed [2, 7–10]. Moreover, various types of ground truth generation from density maps have been proposed, leading to a lack of consensus on which method is best. These generation methods can be categorized into fixed and variable kernel, explained in Section 2.1.

Their framework follows the ground truth with fixed kernel fashion. It is worth mentioning that the number of streams decreased compared to other works [8, 17].

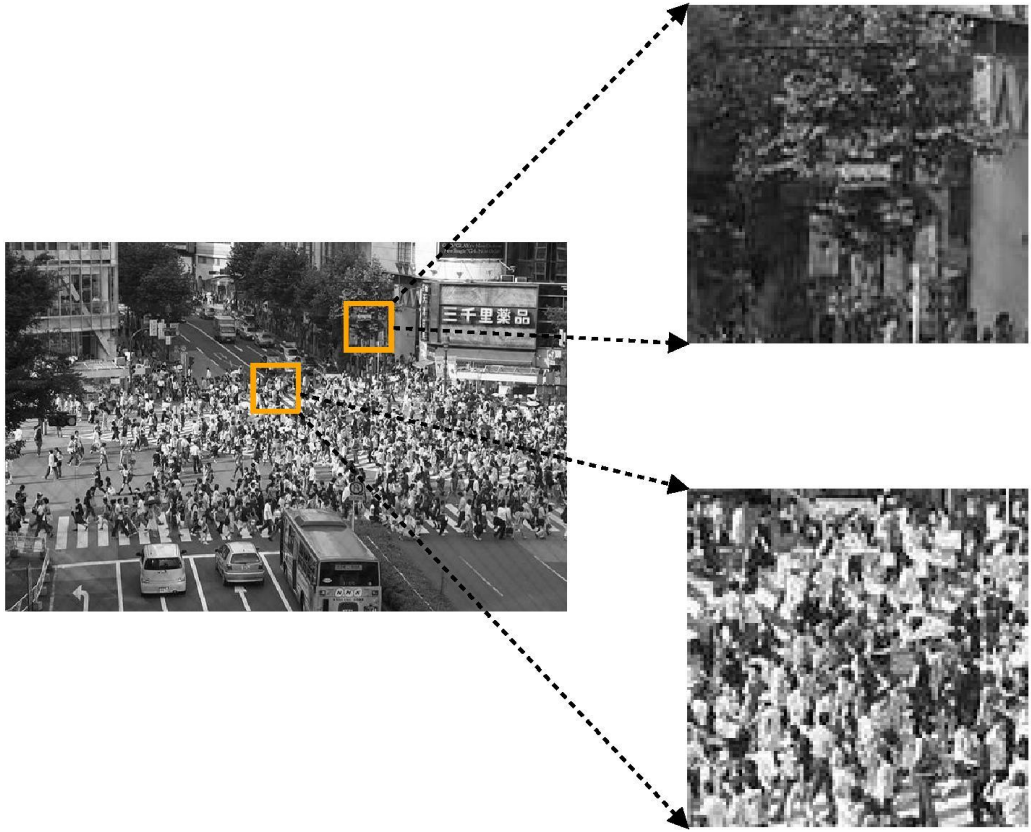

**Fig. 1.** *Illustration of scaling problem in crowd counting. The upper patch is a zoom of the background, whereas the lower patch is a zoom of the people. A challenging task is to differentiate background and people due to overlap and scale of people.*

Onoro et al. [8] proposed a variation of the MSNN, named Hydra CNN (HCNN). This network makes each stream more powerful by stacking more convolutional layers than in MSNN. HCNN is based on the three-stream Counting CNN (CCNN). HCNN learns a mapping between image patches to their corresponding density maps, which differs from the MSNN since this is fully convolutional, such that it can handle random size images. The authors of HCNN designed the network to be scale-aware. Thus, HCNN is fed with a pyramid of patches extracted at multiple scales, where each level of the pyramid is processed by a stream. Then, fully connected layers are used to join information of all the streams. Finally, the prediction is a density map for the patch on top of the pyramid. To define the ground truth, they followed a fixed kernel fashion.

Another way to tackle the scale problem is to improve the network to use various active fields. Thus, Boominathan et al. [7] proposed a deep learning framework with two streams, one with a deep architecture and the other with shallow architecture. The idea behind is that the deep stream was used to capture both high-level semantic (face and body detectors) information, whereas the shallow stream to capture low-level fractures (blob detectors). Finally, they join the streams with a 1x1 convolution and upsample the images using bilinear interpolation, such that the output of the network has the same size as the input. Furthermore, they proposed a multi-scale data augmentation technique to increase the training size.

Sam et al. [9] proposed a three-stream network with a switching module to decide which stream is better for the input images, named switching CNN. Similar to HCNN, it uses patches from the image as input. Then, the stream classifier will choose the best stream to process the patch. Each independent stream is a CNN regressor with different receptive fields and field-of-view, such that it focuses on a specific scale. The granularity of the input patches is important since it is desirable that each patch has a uniform scale distribution. However, this method may create some more specialized streams than others due to unbalanced scale data.

To the best of our knowledge, no previous work based on MSNN has analyzed the effects of the number of streams. Moreover, it is not common practice to evaluate various ground truth generation methods. In this work, we aim to extend our previous work [2] by doing a comprehensive ablation study of the MSNN, specifically by studying the effects of the number of streams. We also evaluate three different methods for density map generation from ground truth, two of them are the most common methods used previously. In addition, we introduce a new method based on face detection and scale interpolation.

## 2. RESEARCH METHOD

In this work, we evaluate (i) the ground-truth construction from people's position and (ii) MSNN

variations with different numbers of streams. These two stages are explained in the following sections.

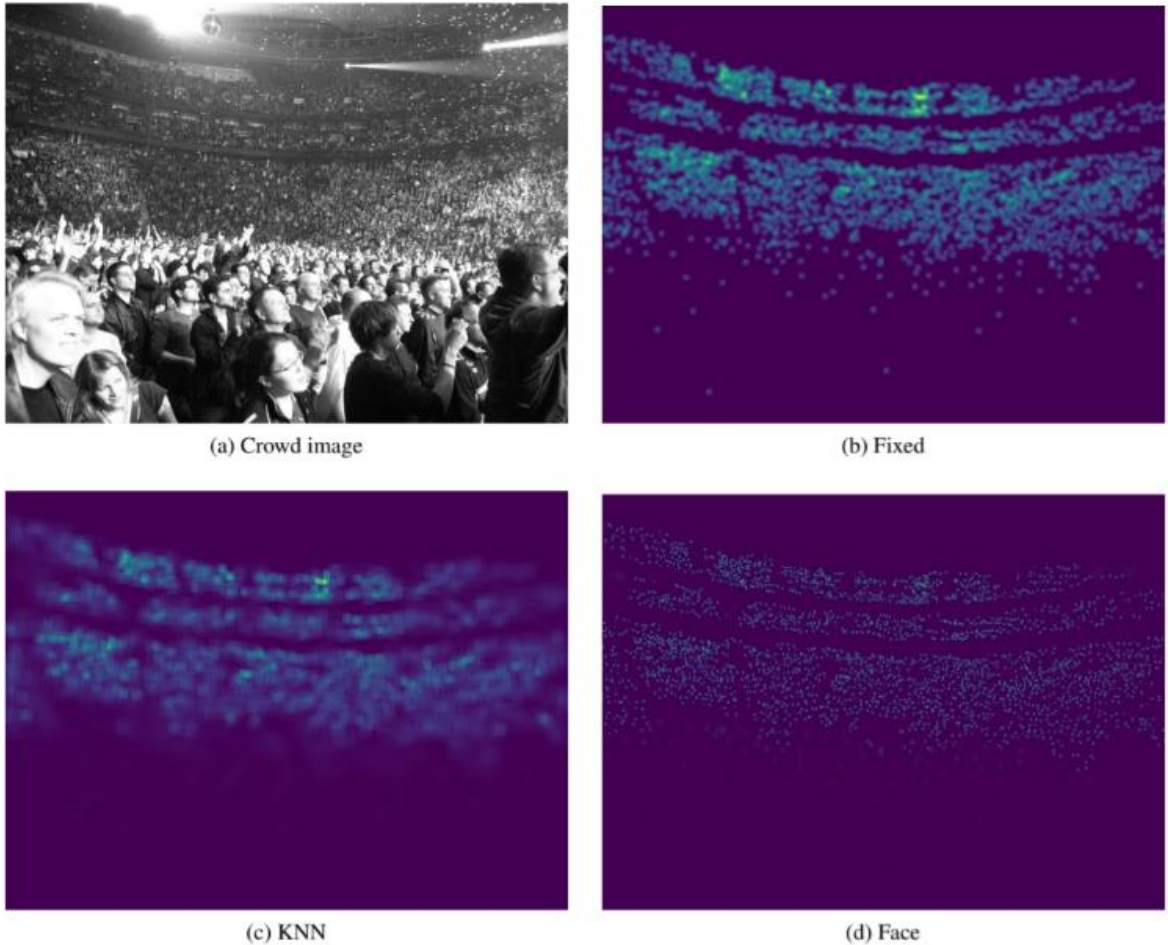

*Fig. 2. Comparison of ground-truth generation using three different methods.*

### 2.1. Ground Truth' Density Map Construction

Crowd counting datasets provide images and positions (usually located in the heads) of each person. Based on these labels, a density map is created since it has been demonstrated [8–10, 17, 18] that such representation is simple, yet effective to predict the number of people present in the scenes when using deep networks. The purpose of the density maps is to describe the density distribution of people in a given image (Figure 2).

Following the work developed by Zhang et al. [17], we assume that $P$ people are located in the image. Given that the $i$-th person is at pixel coordinate $x_i$, for simplicity we use $x_i$ to express both row and column positions. The image, composed of pixel coordinates $x$, is labeled with $P$ heads through the accumulation of many impulse functions, such as:

$$H(x) = \sum_{i=1}^{P} \delta(x - x_i) \quad (1)$$

where the $\delta(.)$ function is defined as:

$$\delta(x - a) = \begin{cases} 1 & if\ x = a, \\ 0 & otherwise. \end{cases} \quad (2)$$

To convert such image to a continuous domain, we convolve it with a Gaussian kernel (with standard deviation $\sigma$) as:

$$F(x) = H(x) * G_\sigma(x) \quad (3)$$

Current literature uses two variations for the size of the Gaussian kernel $\sigma$: (i) a fixed value (for instance, $\sigma = 4$ ) or (ii) a variable value for each person $\sigma_i = \beta d_i$, where $d_i$ is defined as the mean distance to the $k$ closest people and $\beta$ is a regularization parameter.

Authors who employ variable $\sigma$ [17] argue that ground truth created through this way simulates better people's scale such that the network can learn a scale-aware model. On the other hand, authors who employ a fixed $\sigma$ [2] argue that using $k$ closest people introduces errors in poorly crowded regions with small people's scale. Moreover, they showed equal or better results in similar setups.

We consider that both fashions have valid arguments. Thus, we propose a new approach that combines both methods: $\sigma_i$ will be a fixed value for very crowded regions of the image and a variable value otherwise. Let $B = \{b_1, b_2, ..., b_{|B|}\}$ be the list of bounding boxes of detected faces (we used an algorithm for tiny face detection [19] to find them).

Each bounding box $b_i$ is axis-aligned and is defined by a centroid, height, and width. It is expected that the face detection method will not detect all people in the images. Thus, we use $B$ to interpolate missing bounding boxes.

First, for each person $i$ at pixel $x_i$, we must determine if it is inside a crowded region. We initially define an overlap region $r_i$, which is an axis-aligned rectangle centered at $x_i$. To count the number of people around person $i$, due to scale changes, we use a weighted average using the bounding boxes B,

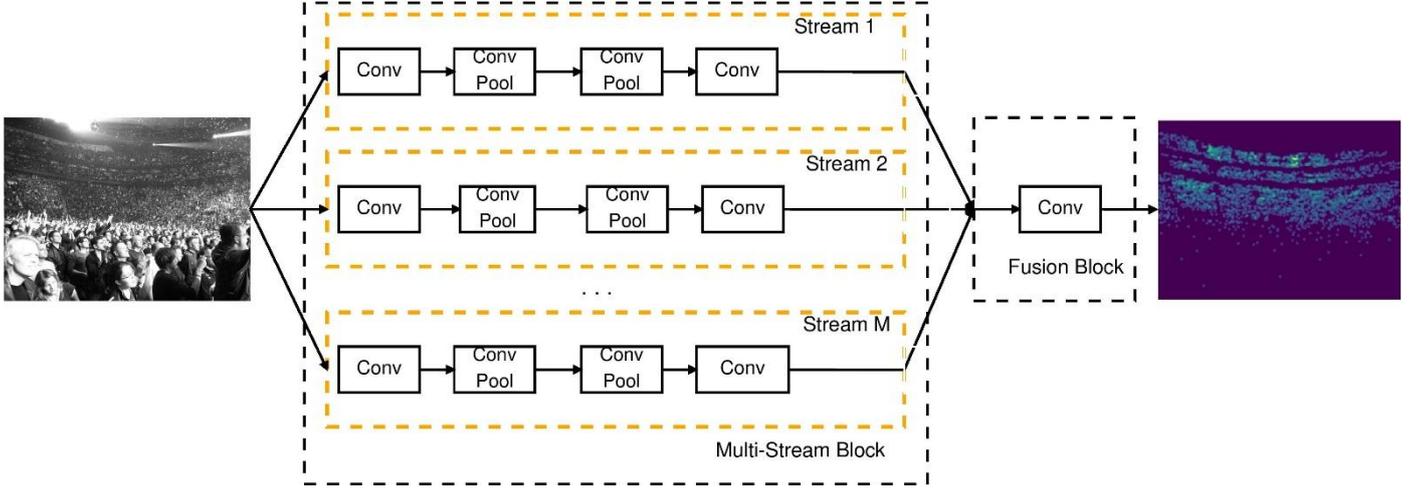

*Fig. 3. Generalization of Multi-Stream Neural Network for crowd counting. Each stream gathers information from different scales. Then, the fusion block merges the information to produce the estimated density map.*

defined as:

$$r_i = \frac{\sum_{j=1}^{|B|} w_{ij}^{overlap} b_j}{\sum_{k=1}^{|B|} w_{ik}^{overlap}} \quad (4)$$

where $w_{ij}^{overlap}$ is defined by the inverse of $\ell_2$ distance between person $i$ and centroid $c_j$ of bounding box $b_j$:

$$w_{ij}^{overlap} = \frac{1}{\parallel x_i - c_j \parallel_2} \quad (5)$$

It is worth mentioning that Equation 4 weights the bounding boxes. We use this form to express that the equation is applied independently for heights and widths of $r_i$ and B.

Second, for every $r_i$, we estimate how many other bounding boxes overlap. If this number is greater than a threshold $T_{overlaps}$, then chances are that person $i$ is at a heavily crowded region. Therefore, we use a fixed predefined size bounding box for it. Otherwise, $x_i$ is not at a heavily crowded region and we can interpolate the bounding boxes with the following weight definition:

$$w_{ij}^{bb} = \frac{1}{\parallel x_i - c_j \parallel_2^{10}} \quad (6)$$

The only difference between Equations 5 and 6 is the power of 10. In the case of Equation 5, we aim to have an accurate estimation of the density context for each person $i$. On the other hand, with Equation 6, we quickly decrease the importance of far elements as they create noise. Thus, bounding box $d_i$ for person $i$ is defined as:

$$d_i = \frac{\sum_{j=1}^{|B|} w_{ij}^{bb} b_j}{\sum_{k=1}^{|B|} w_{ik}^{bb}} \quad (7)$$

Finally, we use height and width of $d_i$ for the Gaussian kernels $\sigma_i = d_i$.

*Table 1 Architecture details for the evaluated MSNN versions.*

|  | $MSNN_1$ | $MSNN_2$ | |
| --- | --- | --- | --- |
| **Multi-Stream Block** | Stream 1 | Stream 1 | Stream 2 |
|  | Conv 3x3x24 | Conv 3x3x24 | Conv 7x7x20 |
|  | Conv 3x3x48 | Conv 3x3x48 | Conv 5x5x40 |
|  | Pool 2x2 | Pool 2x2 | Pool 2x2 |
|  | Conv 3x3x24 | Conv 3x3x24 | Conv 5x5x20 |
|  | Pool 2x2 | Pool 2x2 | Pool 2x2 |
|  | Conv 3x3x12 | Conv 3x3x12 | Conv 5x5x10 |
| **Fusion Block** | Conv 1x1x1 | Conv 1x1x1 | |
|  | | $MSNN_3$ | |
| **Multi-Stream** | Stream 1 | Stream 2 | Stream 3 |
|  | Conv | Conv | Conv |

| Block | 3x3x24 | 7x7x20 | 9x9x20 | |
|---|---|---|---|---|
| | Conv 3x3x48 | Conv 5x5x40 | Conv 7x7x32 | |
| | Pool 2x2 | Pool 2x2 | Pool 2x2 | |
| | Conv 3x3x24 | Conv 5x5x20 | Conv 7x7x16 | |
| | Pool 2x2 | Pool 2x2 | Pool 2x2 | |
| | Conv 3x3x12 | Conv 5x5x10 | Conv 7x7x8 | |
| Fusion Block | Conv 1x1x1 | | | |

$MSNN_4$

| Multi-Stream Block | Stream 1 | Stream 2 | Stream 3 | Stream 4 |
|---|---|---|---|---|
| | Conv 3x3x24 | Conv 7x7x20 | Conv 9x9x20 | Conv 11x11x12 |
| | Conv 3x3x48 | Conv 5x5x40 | Conv 7x7x32 | Conv 9x9x24 |
| | Pool 2x2 | Pool 2x2 | Pool 2x2 | Pool 2x2 |
| | Conv 3x3x24 | Conv 5x5x20 | Conv 7x7x16 | Conv 9x9x12 |
| | Pool 2x2 | Pool 2x2 | Pool 2x2 | Pool 2x2 |
| | Conv 3x3x12 | Conv 5x5x10 | Conv 7x7x8 | Conv 9x9x6 |
| Fusion Block | Conv 1x1x1 | | | |

### 2.2. Multi-Stream Neural Networks

Since the seminal work of Zhang et al. [17], diverse variations of MSNN have been proposed [2, 7–10]. In this work, we generalize the original MSNN and evaluate the number of streams and their behavior with various ground-truth generation methods.

We show a generalization of MSNN architecture in Figure 3. The image is fed to the Multi-Stream Block that has various parallel sequential convolutional layers, named streams. Each stream learns to detect people on a certain scale. Then, the Fusion Block combines the feature maps of each stream to create a final estimation of the crowd.

We propose to evaluate MSNN using one, two, three and four streams. In order to create MSNN with fewer streams, we iteratively remove streams with larger convolutional kernels and change the fusion block according to the new setup (details for each version are shown in Table 1). To train the network, we find the optimal parameters $\theta^*$ (for the network) that minimize the error between the estimated and ground truth density:

$$\theta = arg_\theta \min L(\theta), \quad (8)$$

where the loss function is:

$$L(\theta) = \frac{1}{2|T|} \sum_{i=1}^{|T|} \| \mathcal{F}(X_i, \theta) - F_i \|_2^2, \quad (9)$$

where $\mathcal{F}$ is the function approximated through our network, $\theta$ is a set of learnable parameters in the multi-stream neural network, $X_i$ is the $i$-th input image and $F_i$ its ground-truth density map (Equation 3), $|T|$ is the number of training images, and $\| . \|_2$ is the Euclidean distance.

### 2.3. Training and Data Augmentation

The loss function (Equation 9) is optimized via back-propagation and batch-based stochastic gradient descent. Differently from the work described by Zhang et al. [17], we do not train each stream independently. Due to the two pooling layers, the size of the output is a quarter of the original size, then we resize the ground truth images to compare them with the output. Parameter configuration for training in each crowd counting dataset, *UCF-CC-50* and *ShanghaiTech*, is reported in Table 2.

*Table 2* Hyperparameters for training in each crowd counting dataset.

| | UCF-CC-50 | SHANGHAITECH |
|---|---|---|
| **OPTIMIZER** | Adam | Adam |
| **LEARNING RATE** | 0.00001 | 0.00001 |
| **BATCH SIZE** | 32 | 64 |
| **EPOCHS** | 1000 | 200 |

We perform an extensive data augmentation of the training dataset by creating images with a sliding window of 256×256 pixels and displacement of 70 pixels in each iteration. Further, we add Gaussian and bright/contrast noise. For the *UCF-CC-50* dataset, the augmentation process generates 10032, 10172, 9920, 9724 and 10248 images for five folds, respectively. For the *ShanghaiTech*, the augmentation process generates 65341 and 140801 for part A and B, respectively. Differently from our previous work [2], we kept training and tuning of the hyperparameters simple, as we intended avoid the effects of these factors on the results and have a fair comparison between different network and density map generation setups.

### 3. RESULTS AND DISCUSSION

To evaluate the quality of ground-truth generation methods and MSNN, we use two challenging datasets, summarized as follows.

(1) The *ShanghaiTech* dataset was introduced by Zhan et al. [17]. It was created to encourage research

in crowd counting using deep learning approaches. The dataset has 1198 annotated images with a total of 330164 people with their head positions annotated. It is made up of two parts. Part A is composed of 482 images randomly taken from the Internet, which have different sizes and contain between 501 and 3139 people. There are 300 images for training and 182 for testing. Part B is composed of 716 images taken from a busy street of the metropolitan area of Shanghai, containing between 123 and 578 people. There are 400 images for training and 316 for testing. Unlike other datasets [16], the crowd density varies significantly between the two subsets, making accurate crowd estimation more challenging.

(2) The *UCF-CC-50* dataset was introduced by Idrees et al. [16]. It is a very challenging dataset due to its extreme changes in scale and number of people that varies from 94 to 4543. It contains 50 images extracted from the Internet with different aspect ratios and resolutions. Following the original standard protocol, we report results using a 5-fold cross-validation.

To evaluate the quantitative performance of the MSNN and ground-truth methods, we compute the Mean Absolute Error (MAE) and Root Mean Squared Error (RMSE) metrics, defined as:

$$\text{MAE} = \frac{1}{N}\sum_{i=1}^{N}|y_i - y_i'|, \quad (10)$$

$$\text{RMSE} = \sqrt{\frac{1}{N}\sum_{i=1}^{N}(y_i - y_i')^2}, \quad (11)$$

where *N* is the number of test samples, $y_i$ is the ground-truth count, and $y_i'$ is the estimated count corresponding to the $i$-th sample.

Initially, we make a qualitative assessment of the different methods for density maps generation. Then, we show and analyze results for comparing four different MSNN setups and three ground-truth methods.

### 3.1. Density Maps

We compare the quality of maps generated using fixed kernel, denoted *Fixed*, using variable kernel, denoted *K-NN*, and the proposed hybrid method, denoted *Face*. Figure 2 shows generated maps for an image with large scale variations.

Consider people located far from the camera with a tiny scale. In this case, the *Fixed* method seems to have a good representation, however, it is possible to observe that size of the Gaussian kernel has a significant effect. If a huge value is used, the background will be labeled as people.

Analogously, consider the person on the lower left side of the image. It has a large scale, but the label generated by the Fixed method only considers a tiny part of it, then creating artifacts. The *K-NN* method introduces more artifacts for tiny scales due to the large Gaussian kernel size. This is because it is difficult to find proper hyperparameters $\beta$ and *k* that are suitable for all crowd scenarios.

It is also possible that people with a large scale are almost invisible in the *K-NN* density maps. This same effect appears in the *Face* method. This effect occurs because the Gaussians are normalized to sum one before added to the density map, therefore, larger kernel sizes generate small values. In addition to this effect, the *Face* method is able to have a better estimation of people with tiny and medium scale.

To better understanding the quality of the *Face* method, we analyze the interpolated face scales overlaid on the images (Figure 4). Our proposed ground-truth methods are more general, so it decreases artifacts because it deals with dramatic changes in scales.

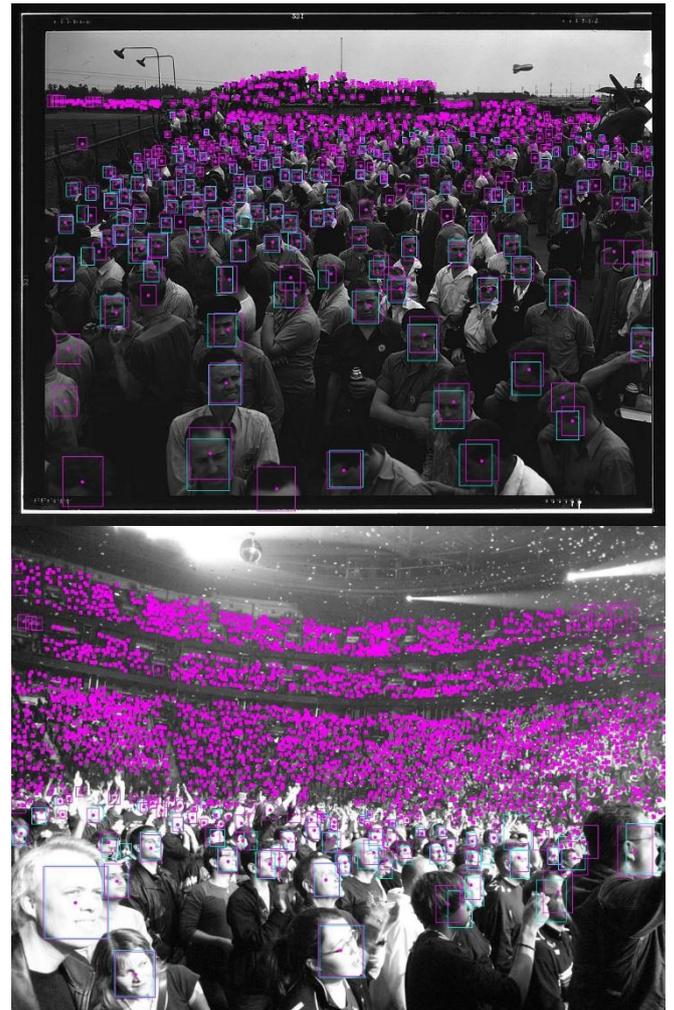

**Figure 4.** Results of face scale interpolation using the proposed algorithm. The bounding boxes detected by the tiny face detection algorithm [19] are shown in cyan, whereas the interpolated bounding boxes are shown in pink.

### 3.2. Crowd Counting

Results for the *UCF-CC-50* dataset are reported in Table 3. It is possible to notice that, for $MSNN_1$, the best results for MAE were obtained with the ground truth generated by the *Face* method and. for RMSE, the *Fixed* method achieved the best results with a difference of 3.06 over the *Face* method. For $MSNN_2$, the *Face* method achieved the best results for MAE and RMSE with a significant difference over the second-best method (for instance, 35.86 for MAE and

16.72 for RMSE). For $MSNN_3$, the *Fixed* method generated the best results for MAE, however, the *Face* method was really close with a tiny difference of 0.05. On the other hand, the *Face* method created the best for RMSE with a difference of 14.23 over the second best. Unexpectedly, *K-NN* method created the best results for $MSNN_4$ with MAE. However, the *Face* method obtained the best results for RMSE. To determine which ground-truth method was the best for *UCF-CC-50*, we averaged the results over the four MSNN setups. It is possible to observe that the *Face* method was superior in terms of MAE and RMSE values.

From the relationship between the number of streams and result quality, the MAE and RMSE values decreased using up to three streams. However, using four streams, it decreased only for MAE and *K-NN* method, whereas it grew in the remaining ones, even for the *Fixed* method, which grew 35.35 and 50.56 for MAE and RMSE, respectively.

Considering the average between the ground-truth generation methods, the best results in MAE and RMSE were obtained with the $MSNN_3$. This may be related to the fact that all the networks have the same simple fusion layer. For MSNN with more streams, the fusion layer must learn more complex functions to map larger tensors to density maps. Overall, the best results for MAE were obtained with $MSNN_4$ and *K-NN* method, whereas, for RMSE, with $MSNN_3$ and the *Face* method.

*Table 3. Results for the proposed multi-stream network (lower scores are better).*

| DATASET | GROUND-TRUTH METHODS | | | | | | | |
|---|---|---|---|---|---|---|---|---|
| UCF-CC-50 | Fixed | | KNN | | Face | | Average | |
| | MAE | RMSE | MAE | RMSE | MAE | RMSE | MAE | RMSE |
| $MSNN_1$ | 517.47 | 729.83 | 539.67 | 741.32 | 514.99 | 732.89 | 524.04 | 734.68 |
| $MSNN_2$ | 429.54 | 652.48 | 421.58 | 605.41 | 385.72 | 588.69 | 412.28 | 615.53 |
| $MSNN_3$ | 373.96 | 568.79 | 398.74 | 590.16 | 374.01 | 554.56 | 382.24 | 571.17 |
| $MSNN_4$ | 409.31 | 619.35 | 368.13 | 614.51 | 379.61 | 556.27 | 385.68 | 596.71 |
| Average | 432.57 | 642.61 | 432.03 | 637.85 | 413.58 | 608.10 | | |
| DATASET | GROUND-TRUTH METHODS | | | | | | | |
| SHANGHAITECH | Fixed | | KNN | | Face | | Average | |
| PART A | MAE | RMSE | MAE | RMSE | MAE | RMSE | MAE | RMSE |
| $MSNN_1$ | 193.37 | 279.95 | 191.64 | 275.23 | 187.56 | 273.02 | 190.86 | 276.07 |
| $MSNN_2$ | 161.77 | 251.33 | 162.85 | 247.99 | 165.20 | 252.34 | 163.27 | 250.55 |
| $MSNN_3$ | 161.61 | 245.39 | 170.45 | 252.16 | 160.80 | 246.40 | 164.29 | 247.98 |
| $MSNN_4$ | 163.26 | 246.31 | 173.95 | 265.09 | 163.38 | 242.66 | 166.86 | 251.35 |
| Average | 170.00 | 255.74 | 174.72 | 260.12 | 169.23 | 253.60 | | |
| DATASET | GROUND-TRUTH METHODS | | | | | | | |
| SHANGHAITECH | Fixed | | KNN | | Face | | Average | |
| PART B | MAE | RMSE | MAE | RMSE | MAE | RMSE | MAE | RMSE |
| $MSNN_1$ | 47.16 | 76.09 | 49.42 | 72.45 | 47.18 | 77.76 | 47.92 | 75.44 |
| $MSNN_2$ | 40.62 | 67.13 | 41.72 | 66.56 | 40.75 | 67.80 | 41.03 | 67.16 |
| $MSNN_3$ | 37.98 | 64.35 | 40.77 | 68.68 | 38.76 | 66.26 | 39.17 | 66.43 |
| $MSNN_4$ | 36.89 | 63.18 | 39.96 | 65.06 | 34.54 | 57.73 | 37.13 | 61.99 |
| Average | 40.66 | 67.69 | 42.97 | 68.19 | 40.31 | 67.39 | | |

Results for the *ShanghaiTech Part A* dataset are reported in Table 3. For $MSNN_1$, the best results were obtained with the *Face* method for MAE and the *Fixed* method for RMSE. For $MSNN_2$, the best results for both MAE and RMSE were obtained with the *Face* method. In this case, the difference with the second-best method was large, 35.86 and 16.72 for MAE and RMSE, respectively. For $MSNN_3$, the best results for MAE were obtained with the *Fixed* method, however, the *Face* method achieved a small difference of 0.05 and the best results for RMSE.

For $MSNN_4$, the best results for MAE were obtained with the *K-NN* method; it achieved a difference of 11.48 with the second-best approach, that is, the *Face* method. Nonetheless, for RMSE, the *Face* method achieved the best results with a difference of 58.24 over the second best. Unlike the results for the *UCF-CC-50* dataset, the best number of streams for MAE is two, but the difference with three streams is 0.99. However, for RMSE, the results followed the same behavior as on the *UCF-CC-50* dataset: the results improved from one to three and achieved worse with four streams.

Considering the average, for *ShanghaiTech Part B* dataset, the best results with one and two streams with MAE are obtained with the *Fixed* method, however, the *Face* method has a small difference of 0.02 and 013, respectively. For RMSE however, *KNN* gives the best results. For three streams, *Fixed* has the best results for MAE and RMSE, and *Face* obtained the second-best. Up to this point, it seems that the *Face* method does not replicate the results shown in the previous datasets, however, with four streams, *Face* is clearly better and, considering the average performance between all network's configurations, it is better but the *Fixed* method has similar results.

Considering the average between the ground truth methods, we can see that the performance improves with the number of streams and, overall, the best results are obtained with four streams and the *Face* method.

## 4. CONCLUSIONS AND FUTURE WORK

In this work, we evaluated the influence of the number of streams in multi-stream networks for the crowd counting problem. Furthermore, we evaluated the two most common strategies for generating ground truth and proposed a new hybrid method based on tiny face detection and scale interpolation.

Extensive experiments demonstrated that using three streams is better on average, however, there are some scenarios where using four streams overcomes other setups. Moreover, experiments show that the use of the proposed hybrid ground-truth generation method is, in fact, better than other widely used schemes.

As directions for future work, we intend to evaluate the creation of synthetic data for training purpose, the generation of higher definition density maps, and more accurate estimations.

**ACKNOWLEDGEMENTS**

The authors are grateful to São Paulo Research Foundation (FAPESP grants #2014/12236-1 and #2016/19947-6) and Brazilian National Council for Scientific and Technological Development (CNPq grants #305169/2015-7, #307425/2017-7 and #309330/2018-1) for their financial support.